\pdfoutput=1
\documentclass[final]{cvpr}

\usepackage{times}
\usepackage{epsfig}
\usepackage{graphicx}
\usepackage{amsmath}
\usepackage{amssymb}

\usepackage[pagebackref=true,breaklinks=true,colorlinks,bookmarks=false]{hyperref}
\newcommand\blfootnote[1]{%
  \begingroup
  \renewcommand\thefootnote{}\footnote{#1}%
  \addtocounter{footnote}{-1}%
  \endgroup
}

\begin{document}

\title{Sim2Real for Self-Supervised Monocular Depth and Segmentation}


\maketitle

\begin{abstract}
Image-based learning methods for autonomous vehicle perception tasks require large quantities of labelled, real data in order to properly train without overfitting, which can often be incredibly costly. While leveraging the power of simulated data can potentially aid in mitigating these costs, networks trained in the simulation domain usually fail to perform adequately when applied to images in the real domain. Recent advances in domain adaptation have indicated that a shared latent space assumption can help to bridge the gap between the simulation and real domains, allowing the transference of the predictive capabilities of a network from the simulation domain to the real domain. We demonstrate that a twin VAE-based architecture with a shared latent space and auxiliary decoders is able to bridge the sim2real gap without requiring any paired, ground-truth data in the real domain. Using only paired, ground-truth data in the simulation domain, this architecture has the potential to generate perception tasks such as depth and segmentation maps. We compare this method to networks trained in a supervised manner to indicate the merit of these results.
\end{abstract}
\blfootnote{
$^1$ University of California, Berkeley

\, \,\,$^2$ Work done as an intern at Ford Greenfield Labs, Palo Alto, USA

\, \,\,$^3$ Ford Greenfield Labs, Palo Alto, USA}
\section{Introduction}
Perception systems are one of the most important systems in autonomous vehicles, allowing them to see and make accurate inferences about their environment. Advances in deep learning have resulted in impressive improvements in accuracy and evaluation time within this field, and have allowed the innovation of autonomous vehicle perception to progress at a very fast rate. A wide variety of deep learning algorithms can be used for perception tasks, and involve usage of LIDAR and image-based methods. However, LIDAR is often quite expensive, and so research has been done into image-based methods as an alternative to LIDAR, being significantly cheaper and quicker to compute, and potentially being able to reach the sophistication and quality in LIDAR cameras.
\begin{figure}[t]
\begin{center}
\fbox{\includegraphics[width=1\linewidth]{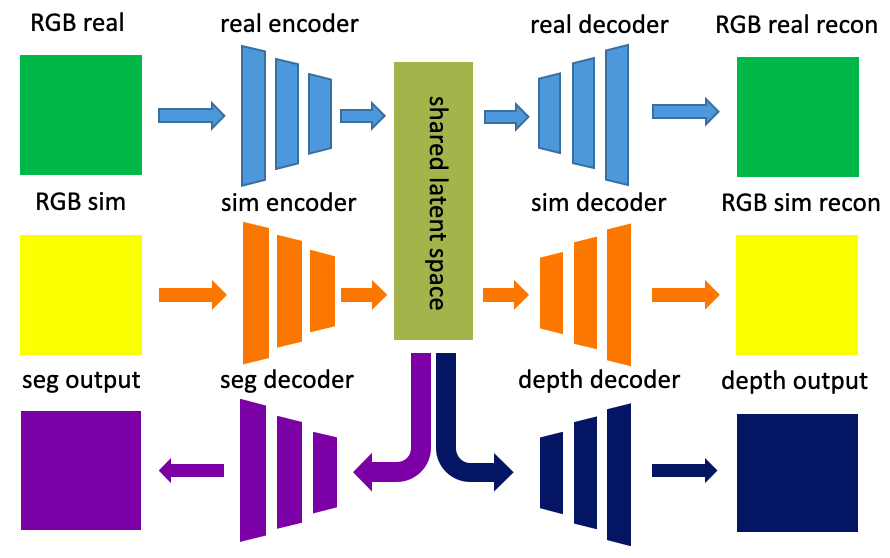}}
\end{center}
   \caption{Our Proposed Architecture, consisting of a twin-VAE structure with a shared latent space, and two auxiliary decoders for depth and segmentation. The network is trained in two steps. The real and sim encoders and decoders form a twin-VAE architecture, and have a shared latent space. For the first step, during training, this shared latent space assumption is used to enforce the latent space embeddings of both RGB real and RGB simulated images to be as close as possible. For the second step, the auxiliary depth and segmentation decoders then take as input these domain-agnostic embeddings to reconstruct depth and segmentation maps.}
\label{arch}
\end{figure}
Examples of important perception tasks include monocular depth estimation and image segmentation. Monocular depth estimation is an ill-posed problem that attempts to ascertain depth information from a single image. Neural networks and machine learning pipelines do this through predicting dense depth maps, or images in which the depth estimate of each pixel of an RGB image is represented by a floating point value. Image segmentation refers to partitioning images using a labelling that identifies a discrete set of object categories. Image segmentation includes semantic segmentation, which performs a pixel-level labelling on the set of object categories (which can include ``tree", ``car", ``signs", among others), as well as instance segmentation, which performs an additional pixel-level shading on each instance of object within the image. Networks predicting these tasks can be trained in an unsupervised, supervised or semi-supervised manner. 

Labelled real data is essential for neural networks trained in a supervised manner. These neural networks require a large amount of input data in order to properly train without overfitting. Gaming engines such as Unity or Unreal Engine are capable of producing large quantities of labelled data at relatively low cost, and so it would be easier and cheaper to train a network on paired simulated-depth map or paired simulated-segmentation map data. The question arises as to whether neural networks trained primarily on paired simulated data would be able to produce accurate depth maps and segmentation maps given real data. Applying such a network directly to real data results in a definitive loss of accuracy, due to the \textit{sim2real gap} that exists between the two domains. 

Work has been done in the field of domain adaptation that can be used to bridge the sim2real gap. In particular, Liu, \textit{et al.} \cite{DBLP:journals/corr/LiuBK17} proposed a twin VAE-GAN architecture, called UNIT, in which the assumption was made that a pair of corresponding images in different domains would result in the same latent representation in a shared latent space. In the event that unpaired data does not exist, cycle-consistency losses can be used to learn a common embedding between the two domains \cite{DBLP:journals/corr/LiuBK17, 8237572, DBLP:journals/corr/ZhuPIE17}. Additionally, \cite{packnet} proposed 3D packing and unpacking layers, which replace traditional striding and pooling with Space2Depth operations and 3D convolutions, which compress key spatial details to allow for the recovery of important spatial information.

We leverage the progress that has been made in these works to produce a neural network architecture designed to bridge the sim2real gap by learning domain invariant features in order to output valid depth and segmentation maps in a domain-agnostic manner. We propose a UNIT-like architecture, wherein the primary component consists of a twin-VAE structure with two encoders, two decoders, and a shared latent space assumption. The encoders encode real and simulated RGB images into the shared latent space, and domain agnostic features from this latent space are decoded back into the real or simulated domains. Additionally, our architecture consists of two external decoders, designed to decode domain agnostic features into valid depth and segmentation maps. Each encoder and decoder makes use of the 3D packing and unpacking layers found in Guizilini, \textit{et al} \cite{packnet}. This allows depth and segmentation maps to be obtained from real images without the assumption of paired ground-truth data in the real domain.

Our architecture is trained in two stages: in the first stage, the twin-VAE structure learns the shared latent space between the simulation and real domains through a combination of cycle-consistency losses and MSE losses; in the second stage, the network is trained with paired simulation-depth and simulation-segmentation data using MSE and cross-entropy losses, respectively. A real image can then be inputted into the network to produce depth and segmentation maps, despite the lack of real-depth and real-segmentation paired data.

Experiments show that while our model performs worse than two state of the art supervised neural networks trained specifically for the depth estimation and segmentation tasks, respectively, it is nevertheless able to recover the general shape of the real ground-truth depth and segmentation maps without having ever seen one, indicating that the idea of domain adaptation from simulation to real for auxiliary perception tasks has merit.


\section{Related Work}
\subsection{Domain Adaptation}
There have been a number of works in the field of unsupervised, semi-supervised and fully-supervised domain adaptation. \cite{10.1145/3400066} provide a good survey of several methods. Recent advances in unsupervised domain adaptation attempt to align the source and target domains by learning domain-invariant features \cite{Zhao2019OnLI}. These methods often include a domain alignment component, such as a classifier, that takes as input these domain invariant features, allowing it to generalize in a domain-agnostic manner. These data specific embeddings can be learned by minimizing the effective distance between two distributions using a divergence measure such as the maximum mean discrepancy loss (MMD), correlation alignment or the Wasserstein metric \cite{Rozantsev2019BeyondSW, cao2018dida, LongC0J15, LongZ0J16, 10.5555/3016100.3016186, Damodaran2018DeepJDOTDJ, Sankaranarayanan2018LearningFS}. 

Other papers approach the concept of domain-invariant feature learning from an adversarial perspective. This often includes a domain discriminator, which attempts to differentiate the source from the target distributions. Sankaranarayanan \textit{et al.} \cite{Sankaranarayanan2018LearningFS} used a discriminator to output one of the four-fold options source-real, source-fake, target-real and target-fake, which is updated each training iteration with an auxiliary classification loss and a within-domain adversarial loss. Sankaranarayanan, \textit{et al.} \cite{Gen2Adapt} utilized a generative adversarial network for the domain-alignment component. \cite{Hoffman_cycada2017, NIPS2017_59b90e10, FuCVPR19-GcGAN} applied CycleGAN \cite{CycleGAN2017} to perform style-transfer for domain adaptation. Liu, \textit{et al.} \cite{DBLP:journals/corr/LiuBK17} used two domain adversarial discriminators to create a twin VAE-GAN architecture, which are used to determine whether the translated images are realistic.

VAE-GANs have been extensively used as networks to perform domain adaptation. VAE-GANs with a shared latent space assumption have been successful in many different areas, including image to image translation and hand pose estimation \cite{DBLP:journals/corr/LiuBK17, huang2018munit, 8099615}. Input images are encoded into the shared latent space using the domain-specific encoder, and are decoded using a separate domain-specific decoder. The image is then encoded and decoded back into the original domain, whereupon a cycle consistency loss is applied.
\subsection{Components}
Weight-sharing is often used to enforce the learning of a joint distribution of images without correspondence supervision \cite{NIPS2016_502e4a16}. Long, et al. \cite{LongC0J15} noted that feature transferability drops significantly in higher layers with increasing domain discrepancy. As a result, many methods use weight-sharing between the two feature extractors in the source and target domains. \cite{NIPS2016_502e4a16} and \cite{DBLP:journals/corr/LiuBK17} perform weight sharing the last few layers to enforce a high-level correspondence between the two domains.

Recent work has also indicated that using losses from auxiliary tasks help regularize the feature embedding, allowing the network to generalize better \cite{Sankaranarayanan2018LearningFS, Jaipuria2020DeflatingDB, Hoffman_cycada2017}. Reconstruction losses are frequently used where paired data is unavailable. The cycle consistency loss, first introduced in \cite{CycleGAN2017}, is used in domain adaptation methods to ensure that local structural information within the images is preserved across domains, most notably in Hoffman, \textit{et al} \cite{Hoffman_cycada2017}, but also in \cite{DBLP:journals/corr/LiuBK17, huang2018munit, 8099615}. Mean squared error reconstruction losses are also widely used as reconstruction losses in VAEs \cite{DBLP:journals/corr/abs-1708-08487}.

In several cases, domain adaptation networks using standard convolutional or transposed convolutional architectures see decreased performances for auxiliary tasks requiring fine-grained representations. Guizilini \textit{et al.} \cite{packnet} propose 3D Packing and Unpacking architectures, which use a Space2Depth operation to pack the spatial dimensions of input feature maps into extra channels. A 3D convolutional layer learns to expand back the compressed spatial features, thereby recovering important spatial information, and is then followed with a standard 2D convolution. 
\begin{figure*}[t]
\begin{center}
\fbox{\includegraphics[width=\textwidth]{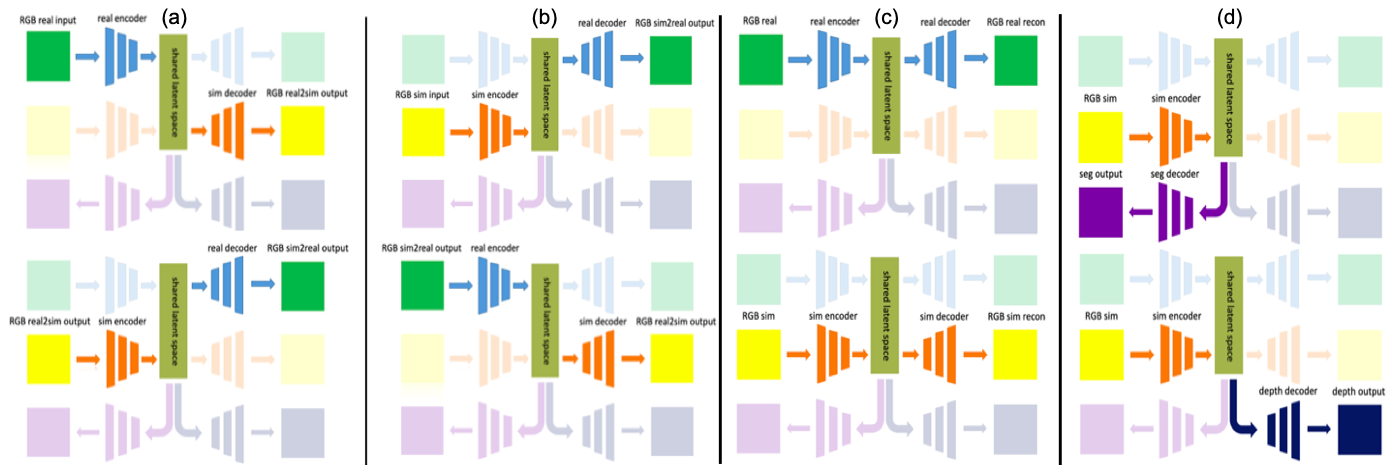}}
\end{center}
   \caption{(a) and (b) describe the images that will be compared using cycle consistency. In (a), the input real image is outputted into the simulation domain through the shared latent space, which is then converted back to the real domain. This reconstructed real image is then compared using cycle consistency with the original real image. In (b), the input simulated image is reconstructed in the same manner. The reconstructed simulated image is then compared using cycle consistency with the original simulated image. 
   (c) and (d) describe the images that will be compared using the MSE losses. In the top half of (c), the input real image is outputted into the real domain through the shared latent space. This reconstructed real image is then compared using MSE with the original real image. In the bottom half of (c), the input simulated image is reconstructed in the same manner. The reconstructed simulated image is then compared using MSE with the original simulated image. In the top half of (d), the output depth maps are compared using paired simulation ground-truth depth data, and finally the output segmentation maps are compared the same way in the bottom half of (d).}
\label{losses}
\end{figure*}
\section{Method}
Our architecture is a UNIT-like architecture based on \cite{DBLP:journals/corr/LiuBK17}. Its main components are a twin-VAE structure consisting of two encoders and two decoders with a shared latent space. These encoders take in  These decoders are classifiers that take as input the domain-agnostic features, and serve as the domain alignment components of our network. This part of the network is trained by itself first before any other part of the network, and the reparameterization trick \cite{kingma:vae} is done during all forward passes in order to backpropagate through the VAE.
\subsection{Twin-VAE architecture}
Let $X_R$ and $X_S$ be the real and simulated domains. In unsupervised domain adaptation, samples $(x_R, x_S)$ are obtained from the two marginal distributions $P_{X_R}(x_R)$ and $P_{X_S}(x_S)$. \cite{DBLP:journals/corr/LiuBK17} notes that any number of joint distributions could yield these marginal distributions, and so enforces a shared latent-space constraint through functions $E_R^*, E_S^*, G_R^*, G_S^*$. Given a pair of corresponding values $(\widetilde{x_R}, \widetilde{x_S})$ from the joint distribution, a shared latent space representation encodes both values in the same manner such that $z = E_R^*(x_R) = E_S^*(x_S)$, and the decoders output images back into their respective domains such that $x_R = G_R^*(z)$ and $x_S = G_R^*(z)$. In practice, to ensure consistency between the learned representations of the two domains within this model, the cycle consistency function is used. 

Let the real and simulation encoders $E_R, E_S$ be parameterized by $\phi,R$ and $\phi,S$ respectively. Similarly let the real and simulation decoders $G_R, G_S$ be parameterized by $\theta,R$ and $\theta,S$ respectively. Then an encoder approximates the  variational distribution $q(x)$ with the distribution $q_{\phi}(z|x)$, and a decoder learns the decoding distribution $p_{\theta}(x|z)$. Let the prior over the latent variables be a centered, isotropic, multivariate Gaussian. We want to maximize $p_\theta(x|z)$, and therefore jointly optimize the expected value $\mathbb{E}_{q_{\phi(z|x)}}[\log p_\theta(x|z)]$. We also intend to minimize the Kullback–Leibler (KL) divergence such that the latent distribution $q(z|x)$ is as close as possible to a zero-mean, unit-variance normal distribution $N$ with PDF $p_{N} = \frac{1}{(2\pi)^{d/2}}\exp(-\Vert x \Vert_2^2/2)$. This leads to the general VAE loss function
\begin{align}
\begin{aligned}
\mathcal{L}_\mathrm{VAE} = d_{\mathrm{KL}}(q_{\phi}(z|x) \parallel p_N(z)) - \mathbb{E}_{z\sim q_{\phi(z|x)}}[\log p_G(x|z)]
\end{aligned}
\end{align}
whereupon we derive two of our reconstruction losses: 
\begin{align}
\begin{aligned}
\mathcal{L}_{R,\mathrm{KL}} = d_{\mathrm{KL}}(q_{\phi,R}(z_R|x_R) \parallel p_{N}(z_R)) \\ - \mathbb{E}_{z_R\sim q_{\phi,R(z_R|x_R)}}[\log p_{G_R}(x_R|z_R)] 
 \end{aligned}\\
 \begin{aligned}
\mathcal{L}_{S,\mathrm{KL}} = d_{\mathrm{KL}}(q_{\phi,S}(z_S|x_S) \parallel p_{N}(z_S)) \\ - \mathbb{E}_{z_S\sim q_{\phi,S(z_S|x_S)}}[\log p_{G_S}(x_S|z_S)] 
 \end{aligned}
\end{align}
We then make use of two additional MSE losses to maximize the quality of the reconstruction of each decoder:
\begin{align}
\mathcal{L}_{R,\mathrm{MSE}} &= \lVert x_R - (G_R \circ E_R)(x_R) \rVert_2 \\
\mathcal{L}_{S,\mathrm{MSE}} &= \lVert x_S - (G_S \circ E_S)(x_S) \rVert_2
\end{align}
We further use MMD losses to maximize the mutual information between each distribution and the standard multivariate normal distribution, based on the kernel $k(x, y) = \exp(-\lVert x - y \rVert_2^2/(2\sigma ^2))$. Following \cite{Rustamov2019ClosedformEF}, we have:
\begin{align}
    \begin{aligned}
        \mathcal{L}_{R,\mathrm{MMD}} = \mathbb{E}_{a,\tilde{a}\sim N}[k(a, \tilde{a})] \hspace{50pt}\\- 2\mathbb{E}_{\tilde{a}\sim N, \tilde{z}_R\sim q_{\phi,R}}[k(\tilde{a}, \tilde{z}_R)] \hspace{15pt}\\+ \mathbb{E}_{z_R, \tilde{z}_R\sim q_{\phi,R}}[k(z_R, \tilde{z_R})]\hspace{22pt}
    \end{aligned}\\
    \begin{aligned}
        \mathcal{L}_{S,\mathrm{MMD}} = \mathbb{E}_{a,\tilde{a}\sim N}[k(a, \tilde{a})] \hspace{50pt}\\- 2\mathbb{E}_{\tilde{a}\sim N, \tilde{z}_S\sim q_{\phi,S}}[k(\tilde{a}, \tilde{z}_S)] \hspace{16pt}\\+ \mathbb{E}_{z_S, \tilde{z}_S\sim q_{\phi,S}}[k(z_S, \tilde{z_S})]\hspace{24pt}
    \end{aligned}
\end{align}
In practice we sample $a$ from such a Gaussian distribution, and let $a = \tilde{a}$, and $z_R=\tilde{z}_R$. In accordance with \cite{DBLP:journals/corr/LiuBK17} we then use cycle-consistency losses between each input sample and the cross-domain reconstructed sample to ensure that the local structural information is preserved:
\begin{align}
    \begin{aligned}
        \mathcal{L}_{R,\mathrm{CC}} = \lVert (G_R \circ E_S \circ G_S \circ E_R)(x_R) - x_R \lVert_1
    \end{aligned} \\
    \begin{aligned}
        \mathcal{L}_{S,\mathrm{CC}} = \lVert (G_S \circ E_R \circ G_R \circ E_S)(x_S) - x_S \lVert_1
    \end{aligned} 
\end{align}
Our total loss function, minimized over $E_R, E_S, G_R$ and $G_S$ is the sum of these eight functions:
\begin{align}
    \begin{aligned}
        \mathcal{L} = \mathcal{L}_{R,\mathrm{KL}} + \mathcal{L}_{S,\mathrm{KL}} + \mathcal{L}_{R,\mathrm{MSE}} + \mathcal{L}_{S,\mathrm{MSE}} \hspace{50pt} \\ +\mathcal{L}_{R,\mathrm{MMD}} + \mathcal{L}_{S,\mathrm{MMD}} + \mathcal{L}_{R,\mathrm{CC}} +  \mathcal{L}_{S,\mathrm{CC}} \hspace{30pt}
    \end{aligned}
\end{align}
\subsection{Components}
The last three weights of each encoder are shared with each other, and the first three weights of each decoder are shared with each other. Before training, we perform a Gaussian weight initialization. We incorporate the 3D packing and unpacking layers proposed by Guizilini \textit{et al.} in our architecture, and perform a 2D convolution followed by the InstanceNorm and ReLU operations after the packing and 3D convolution operations. Each encoder $E_i$ consists of six of these packing layers as well as other convolutional layers in the manner described in the appendix. Each decoder $D_i$ consists of six unpacking layers in a manner opposite to that of each encoder. 

This part of the network is trained first, before the auxiliary decoders, and takes unpaired real and simulation images as input. Unlike Liu, \textit{et al.} and similar works, we do not make use of a discriminator because during this step, we only intend to make the embeddings agnostic to both simulation and real, and are thus uninterested in the quality of the sim2real conversions. Additionally, recent work has shown that the losses incorporated from auxiliary tasks (depth and segmentation) serve to regularize the embeddings \cite{Jaipuria2020DeflatingDB, Hoffman_cycada2017, Sankaranarayanan2018LearningFS}. Figure 2 and Figure 3 provide illustrations as to what images our network compares for the MSE and cycle consistency losses, respectively.
\subsection{Auxiliary Decoders}
Our architecture contains additional decoders corresponding to each auxiliary perception task that take domain-agnostic embeddings from the latent space and output images. In our architecture, there are two such decoders that perform monocular depth estimation and semantic segmentation. The auxiliary decoders are not identical to the simulation and real decoders, as they do not share any decoder weights. This parts of the network is trained after the twin-VAE using additional losses. Let $G_D$ and $G_I$ refer to the depth and semantic segmentation decoders, respectively. 

\subsubsection{Monocular Depth Estimation}
For monocular depth estimation, similar to the first section, we minimize the KL divergence so that the latent distribution $q(z|x)$ is as close as possible to a zero-mean, unit-variance normal distribution $N$. In order to maintain the domain-agnostic nature of the latent space, we do this twice: between $N$ and the simulation encoder, and between $N$ and the real encoder. Let $x_R' = G_R(z_S)$ and $z_R' = E_R(x_R')$. Then we have that
\begin{align}
\begin{aligned}
\mathcal{L}_{R\rightarrow D,\mathrm{KL}} = d_{\mathrm{KL}}(q_{\phi,R}(z'_R|x'_R) \parallel p_{N}(z'_R)) \\ - \mathbb{E}_{z'_R\sim q_{\phi,R}}[\log p_{G_R}(x'_R|z'_R)] 
 \end{aligned}\\
 \begin{aligned}
\mathcal{L}_{S\rightarrow D,\mathrm{KL}} = d_{\mathrm{KL}}(q_{\phi,S}(z_S|x_S) \parallel p_{N}(z_S)) \\ - \mathbb{E}_{z_S\sim q_{\phi,S}}[\log p_{G_S}(x_S|z_S)] 
 \end{aligned}
\end{align}
We additionally maximize the mutual information using the MMD loss between $z_S$ and $z_R$ using the Gaussian kernel in the following form. Given some (unpaired) real data $x_R$ with latent representation $z_R$, let $x_S' = G_S(z_R)$ and $z_S' = E_S(x_S')$. Then we have that
\begin{align}
    \begin{aligned}
        \mathcal{L}_{S\rightarrow R,\mathrm{MMD}} = \mathbb{E}_{z_S,\tilde{z}_S\sim q_{\phi,S}}[k(z_S, \tilde{z}_S)] \hspace{50pt}\\- 2\mathbb{E}_{\tilde{z}_S\sim q_{\phi,S}, \tilde{z}_R'\sim q_{\phi,R}}[k(\tilde{z}_S, \tilde{z}_R')] \hspace{25pt}\\+ \mathbb{E}_{z_R', \tilde{z}_R'\sim q_{\phi,R}}[k(z_R', \tilde{z}_R')]\hspace{50pt}
    \end{aligned}\\
    \begin{aligned}
        \mathcal{L}_{R\rightarrow S,\mathrm{MMD}} = \mathbb{E}_{z_S',\tilde{z}_S'\sim q_{\phi,S}}[k(z_S', \tilde{z}_S')] \hspace{50pt}\\- 2\mathbb{E}_{\tilde{z}_S'\sim q_{\phi,S}, \tilde{z}_R\sim q_{\phi,R}}[k(\tilde{z}_S', \tilde{z}_R)] \hspace{25pt}\\+ \mathbb{E}_{z_R, \tilde{z}_R\sim q_{\phi,R}}[k(z_R, \tilde{z}_R)]\hspace{50pt}
    \end{aligned}
\end{align}
As before, we let $z_i = \tilde{z}_i$ and $z_i' = \tilde{z}_i'$ in practice. And finally we use the ground-truth paired simulated depth data $d_T$ with the MSE loss:
\begin{align}
\mathcal{L}_{R,\mathrm{MSE}} &= \lVert d_T - (G_D \circ E_S)(x_S) \rVert_2 \\
\mathcal{L}_{S,\mathrm{MSE}} &= \lVert d_T - (G_D \circ E_R)(x_R') \rVert_2
\end{align}
Our final loss is comprised of these six elements:
\begin{align}
    \begin{aligned}
        \mathcal{L}_\mathrm{depth} = \mathcal{L}_{R\rightarrow D,\mathrm{KL}} + \mathcal{L}_{S\rightarrow D,\mathrm{KL}} + \mathcal{L}_{S\rightarrow R,\mathrm{MMD}} \hspace{20pt} \\ + \mathcal{L}_{R\rightarrow S,\mathrm{MMD}} + \mathcal{L}_{R,\mathrm{MSE}} + \mathcal{L}_{S,\mathrm{MSE}} \hspace{20pt}
    \end{aligned}
\end{align}
\begin{figure*}[t]
\begin{center}
\fbox{
\includegraphics[width=0.5\textwidth]{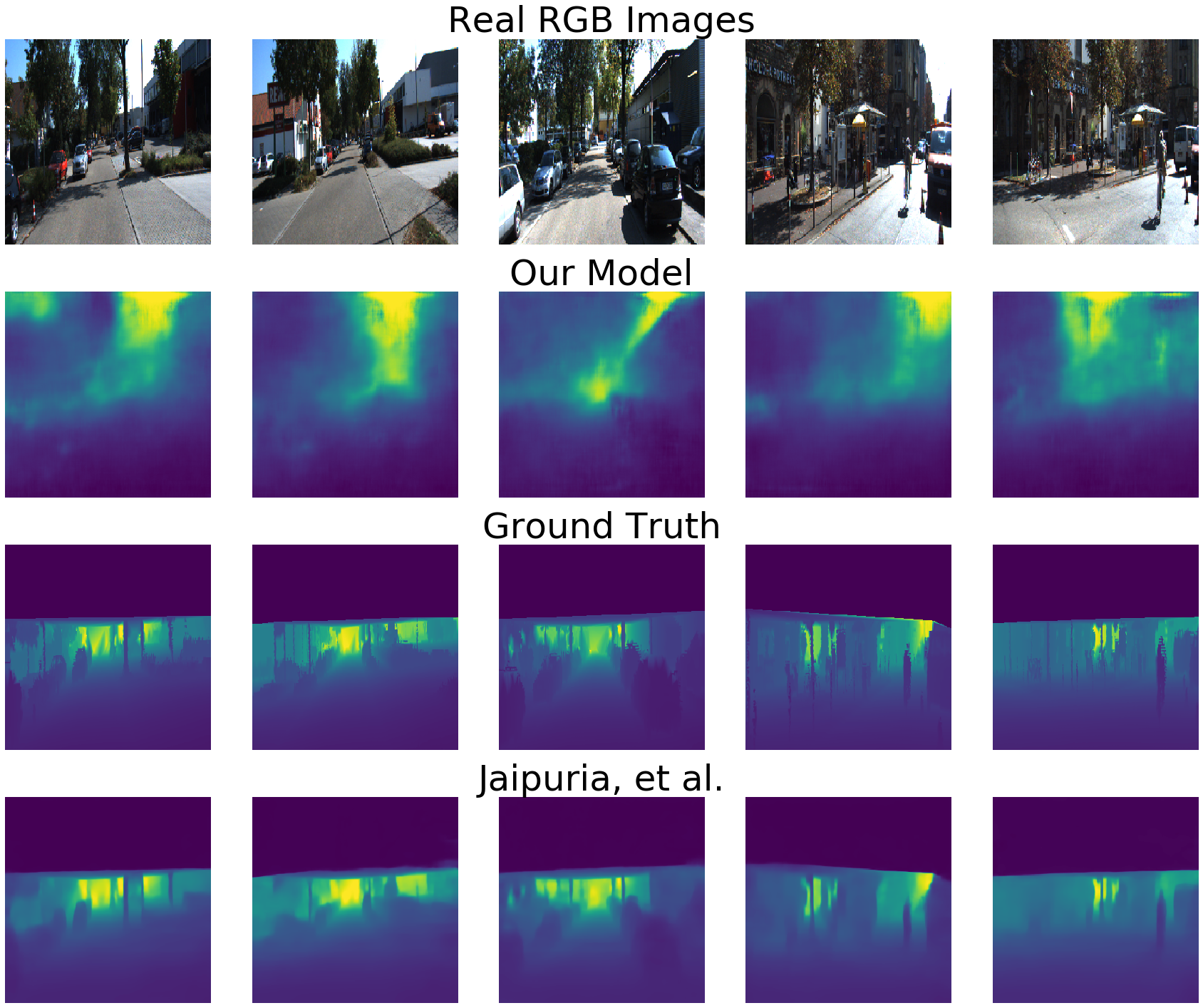}
\includegraphics[width=0.5\textwidth]{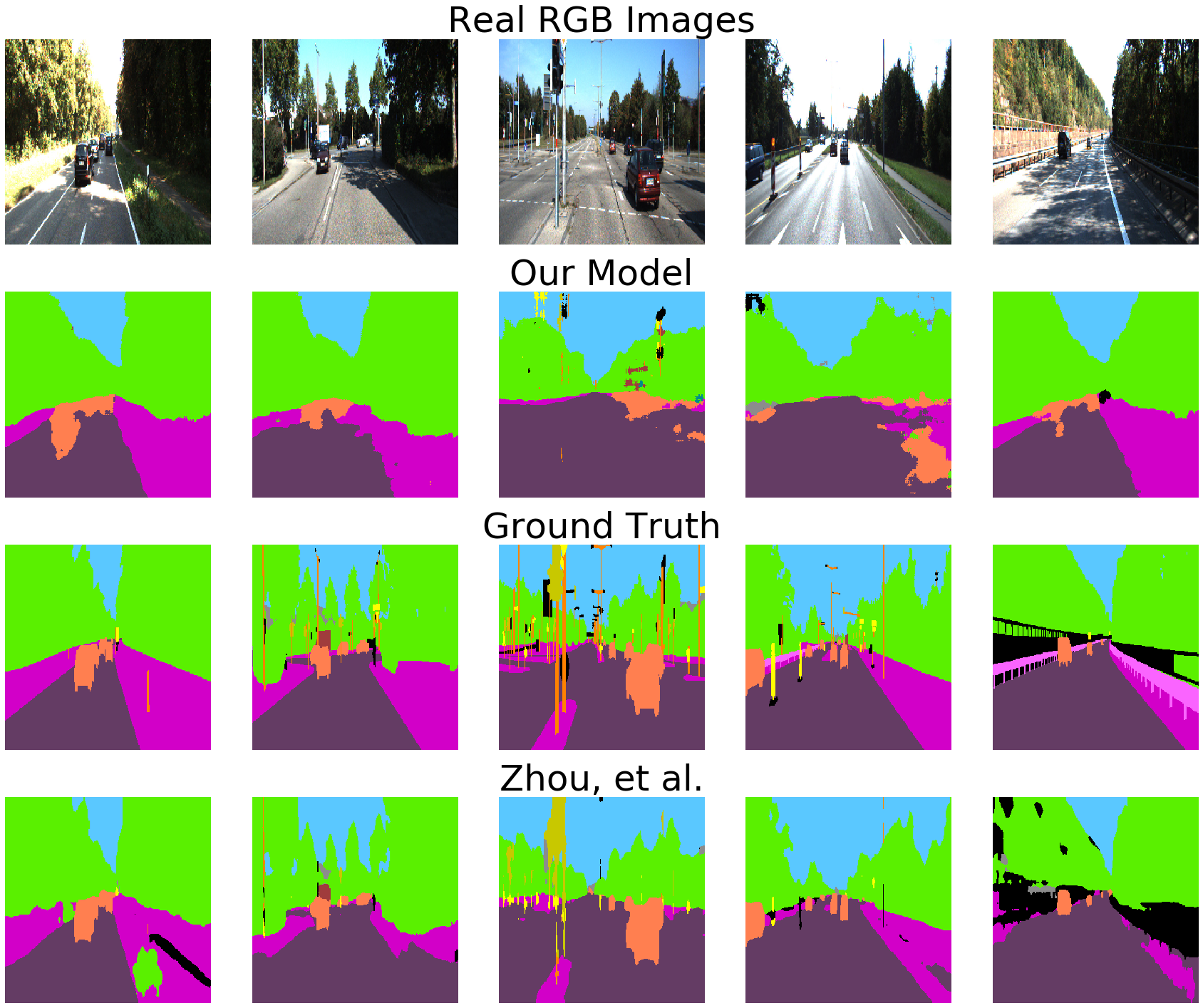}
}
\end{center}
   \caption{Examples of Results for Depth Estimation and Semantic Segmentation. These results indicate that our method recovers the general shape of the ground truth in both tasks, despite being trained only on simulated data. Jaipuria, \textit{et al.} and Zhou, \textit{et al.} perform better, but they are directly supervised on real ground-truth depth and segmentation data, unlike our method, which only sees these images in the simulated domain.}
\label{depthsem}
\end{figure*}
\subsubsection{Semantic Segmentation}
Most of the mathematics for semantic segmentation is the same as for monocular depth estimation. However, instead of an MSE reconstruction loss, we use a pixel-wise cross entropy loss. Let $c$ be the class, and $C$ the total number of classes. The semantic segmentation decoder outputs a prediction of raw logits of size $C$x$256$x$256$. Then let $x[i]$ refer to indexing the $i$th element of $x$. We have that: 
\begin{align}
    \begin{aligned}
        \mathcal{L}_{S, \mathrm{CE}} = -\sum_{c}^C\log \left(\frac{ \exp((G_I \circ E_S)(x_S)[c])}{\sum_{j} \exp((G_I \circ E_S)(x_S)[j])}\right)
    \end{aligned}\\
    \begin{aligned}
        \mathcal{L}_{R, \mathrm{CE}} = -\sum_{c}^C\log \left(\frac{ \exp((G_I \circ E_R)(x_R')[c])}{\sum_{j} \exp((G_I \circ E_R)(x_R')[j])}\right)
    \end{aligned}
\end{align}
Our final loss is comprised of the same KLD and MMD losses as the depth task, with these two cross-entropy losses:
\begin{align}
    \begin{aligned}
        \mathcal{L}_\mathrm{seg} = \mathcal{L}_{R\rightarrow D,\mathrm{KL}} + \mathcal{L}_{S\rightarrow D,\mathrm{KL}} + \mathcal{L}_{S\rightarrow R,\mathrm{MMD}} \\\hspace{20pt}+ \mathcal{L}_{R\rightarrow S,\mathrm{MMD}} + \mathcal{L}_{R,\mathrm{CE}} + \mathcal{L}_{S,\mathrm{CE}}\hspace{20pt}
    \end{aligned}
\end{align}
\section{Experiments}
\subsection{Datasets}
For our experiments, we trained our network using the KITTI \cite{Geiger2012CVPR, Geiger2013IJRR} and VKITTI \cite{cabon2020vkitti2, gaidon2016virtual} datasets. The KITTI dataset consists of images taken with cameras in the real world, and is frequently used for depth evaluation. The VKITTI dataset contains simulated scenes taken in the day, at night, and under different weather conditions such as fog and rain. The KITTI dataset contains ground-truth sparse depth maps and semantic segmentation maps. The VKITTI dataset contains ground-truth dense depth maps and semantic segmentation maps. When training our network, we used the dense depth and semantic segmentation maps from the VKITTI dataset in addition to real-domain images from the KITTI dataset. We used two supervised models for comparison purposes. Zhou \textit{et al.} \cite{zhou2017scene, zhou2018semantic} provided a pre-trained model (ResNet50dilated + PPM Deepsup) trained on the ADE20K dataset for semantic segmentation. Jaipuria \textit{et al.} \cite{Jaipuria2020DeflatingDB} provided a model for depth estimation, trained on a 90-10 split of the KITTI dataset. 

\subsection{Implementation Details}
The network was written using the PyTorch library, based off the code for Chakravarty, \textit{et al} \cite{8793530}. We borrowed code for the 3D Packing and Unpacking layers from Guizilini \textit{et al} \cite{packnet}. We use the Adam optimizer \cite{kingma2014method} with $(\beta_1, \beta_2) = 0.9, 0.999$ and a learning rate of 0.0001. The first section of the network, the sim2real twin-VAE, was trained with 2126 images from the VKITTI dataset and 2126 images from the KITTI dataset for around 300 epochs. The second section of the network, with the depth and segmentation tasks, was trained with 21,260 images from the VKITTI dataset for around 25 epochs.
\begin{table}
\footnotesize 
\begin{center}
\textbf{Depth}
\vspace{1pt}
\resizebox{\columnwidth}{!}{%
\begin{tabular}{|p{2cm}|p{1.2cm}|p{1.2cm}|p{1.2cm}|p{1.2cm}|} \hline 
Network & Abs. Rel. & Sq. Rel. & RMSE & log RMSE \\ \hline
Jaipuria, \textit{et al.} & 0.148 & 0.806 & 4.05 & 0.329 \\ \hline
Ours & 0.752 & 20.7 & 16.1 & 0.580 \\ \hline
\end{tabular}}
\vspace{1pt}
\textbf{Semantic Segmentation}
\vspace{1pt}
\resizebox{\columnwidth}{!}{%
\begin{tabular}{|p{2cm}|p{2.5cm}|p{2.5cm}|} \hline 
Network & mIOU & Accuracy \\ \hline
Zhou \textit{et al.} & 0.128 & 86.2\% \\ \hline
Ours & 0.0630 & 54.9\% \\ \hline
\end{tabular}}
\end{center}
\caption{A comparison of results of our architecture with two supervised models for depth and semantic segmentation. For all of the depth metrics, lower is better, and for the segmentation metrics, higher is better.}
\end{table}

Code written by Ford for \cite{Jaipuria2020DeflatingDB} was used to generate .png images of the sparse depth maps provided by the KITTI dataset, which were used as ground truth images. The VKITTI dataset ground truth for semantic segmentation consists of 14 classes, which differs in number from CityScapes and other datasets. In order to handle the different numbers of classes, we identified 12 classes common to the VKITTI, CityScapes and KITTI datasets (terrain, sky, vegetation, building, road, guard rail, traffic sign, traffic light, pole, truck, car, van). After predicting the total number of classes using each network, we then isolated these twelve classes to compare to the ground truth.

To obtain metrics for segmentation, we compared our model with \cite{zhou2018semantic} on a selection of 200 images of the KITTI dataset with provided ground-truth segmentation data. For this, we used code in the GitHub repository provided with \cite{zhou2017scene, zhou2018semantic} \footnote[4]{Code can be found at \url{https://github.com/CSAILVision/semantic-segmentation-pytorch}}. For depth, we compared our model with Jaipuria \textit{et al.} \cite{Jaipuria2020DeflatingDB} on the testing part of a 90-10 split of the KITTI dataset, or 100 images of the KITTI dataset. The model from \cite{Jaipuria2020DeflatingDB} was trained for 151 epochs on a selection of 4500 images.

We use the absolute relative difference, square relative difference, root mean squared error and log root mean squared error as metrics for verification of depth accuracy. We use mean intersection over union and pixel accuracy for semantic segmentation. All images (predicted, ground truth) were resized to 256x256 before comparison. The averaged results are summarized in Table 1.
\subsection{Results}
Despite our model predictably performing worse than supervised networks trained for depth and semantic segmentation, the results (both qualitative and quantitative) indicate that there is potential for sim2real transfer to occur based on our architecture with further work. There are several images for segmentation that are able to individually obtain 75\% pixel accuracy, and there are several images for depth that are individually able to obtain 10 RMSE. 
\section{Future Work}
We estimate that much benefit could be obtained by training the first section of the twin-VAE on the full dataset of 21,260 images, rather than on just 2126 images. Additionally, we intend to experiment with task-based losses such as the Depth Smoothness Loss found in \cite{jadon2020comparative, jadon2020survey}. Based on related work, we intend to experiment further with the training stages, specifically seeking to obtain a comparison between training the entire network at once and training it in separate stages as we have done above. We will also consider mixing additional synthetic datasets during training in accordance with the work done in \cite{Ranftl2020} to improve on generalizability of the network, and conduct ablation studies.

\section{Conclusion}
We introduce a neural network architecture, a twin VAE-based architecture with a shared latent space and auxiliary decoders, to mitigate the cost of obtaining real-world data for autonomous vehicle perception tasks by performing domain adaptation from the simulated domain to the real domain. This architecture shows promise in being able to generate perception tasks such as depth and segmentation maps without requiring any ground-truth real paired data in these domains. The network is trained with the shared latent space assumption, letting depth and segmentation networks train on domain-agnostic embeddings. Experiments have shown that this type of architecture has promise, and future work will focus on improving several bottlenecks and issues to increase the performance and generalization capabilities of this network.

{\small
\bibliographystyle{ieee_fullname}
\bibliography{cvpr}
}
\clearpage
\section*{Appendix}
\subsection*{Model Architecture}

Please see Table 2 for more information on the network architectures for our encoders and decoders.

\begin{table}[!h]
\footnotesize 
\begin{center}
\resizebox{\columnwidth}{!}{%
\begin{tabular}{|p{0.1cm}|p{6cm}|p{1.6cm}|p{0.1cm}|} \hline
Id. & Description & Output & K \\
\hline
\end{tabular}}
\vspace{1pt}
\textbf{Input}
\vspace{1pt}
\resizebox{\columnwidth}{!}{%
\begin{tabular}{|p{0.1cm}|p{6cm}|p{1.6cm}|p{0.1cm}|} \hline 
0 & Input RGB image & 3x256x256 & \, \\ \hline
\end{tabular}}
\vspace{1pt}
\textbf{Encoder}
\vspace{1pt}
\resizebox{\columnwidth}{!}{%
\begin{tabular}{|p{0.1cm}|p{6cm}|p{1.6cm}|p{0.1cm}|} \hline 
1 & Conv2D + LeakyReLU & 64x256x256 & 7 \\ \hline
2 & Packing + Conv2D + InstanceNorm & 76x128x128 & 3 \\ \hline
3 & Packing + Conv2D + InstanceNorm & 88x64x64 & 3 \\ \hline
4 & Packing + Conv2D + InstanceNorm & 100x32x32 & 3 \\ \hline
5 & Packing + Conv2D + InstanceNorm & 128x16x16 & 3 \\ \hline
6 & Packing + Conv2D + InstanceNorm & 200x8x8 & 3 \\ \hline
7 & Conv2D + InstanceNorm + Conv2D + InstanceNorm & 200x8x8 & 3 \\ \hline
8 & Conv2D + InstanceNorm + Conv2D + InstanceNorm & 200x8x8 & 3 \\ \hline
9 & Packing + Conv2D + InstanceNorm & 250x4x4 & 3 \\ \hline
\end{tabular}}
\vspace{1pt}
\textbf{Shared Encoder}
\vspace{1pt}
\resizebox{\columnwidth}{!}{%
\begin{tabular}{|p{0.1cm}|p{6cm}|p{1.6cm}|p{0.1cm}|} \hline 
10 & Conv2D + BatchNorm & 300x4x4 & 3 \\ \hline
\end{tabular}}
\vspace{1pt}
\textbf{Shared Decoder}
\vspace{1pt}
\resizebox{\columnwidth}{!}{%
\begin{tabular}{|p{0.1cm}|p{6cm}|p{1.6cm}|p{0.1cm}|} \hline 
11 & ConvTranspose2D + BatchNorm (*) & 250x4x4 & 3 \\ \hline
\end{tabular}}
\vspace{1pt}
\textbf{Decoder}
\vspace{1pt}
\resizebox{\columnwidth}{!}{%
\begin{tabular}{|p{0.1cm}|p{6cm}|p{1.6cm}|p{0.1cm}|} \hline 
12 & Conv2D + InstanceNorm + Unpacking & 200x8x8 & 3 \\ \hline
13 & Conv2D + InstanceNorm + Conv2D + InstanceNorm & 200x8x8 & 3 \\ \hline
14 & Conv2D + InstanceNorm + Conv2D + InstanceNorm & 200x8x8 & 3 \\ \hline
15 & Conv2D + InstanceNorm + Unpacking & 128x16x16 & 3 \\ \hline
16 & Conv2D + InstanceNorm + Unpacking & 100x32x32 & 3 \\ \hline
17 & Conv2D + InstanceNorm + Unpacking & 88x64x64 & 3 \\ \hline
18 & Conv2D + InstanceNorm + Unpacking & 76x128x128 & 3 \\ \hline
19 & Conv2D + InstanceNorm + Unpacking & 64x256x256 & 3 \\ \hline
20 & ConvTranspose2D + Tanh (**) & 3x256x256 & 7 \\ \hline

\end{tabular}}
\end{center}
\caption{A description of our architecture. Unless otherwise specified, all activations are ReLU. These layers are identical for each encoder and each decoder, with one exception. (*) For the auxiliary decoders, the layers listed in "Shared Decoder" are not shared with each other, and so are just part of the decoder. (**) There is no tanh layer in the semantic segmentation decoder, and the output of the ConvTranspose2D is of size 15x256x256.}
\end{table}
\end{document}